\documentclass[runningheads]{llncs}
\usepackage{graphicx}

\usepackage{comment}
\usepackage{amsmath,amssymb} 
\usepackage{color}

\usepackage{booktabs}
\usepackage[caption=false]{subfig}
\usepackage{dsfont}
\usepackage{makecell}
\usepackage{floatrow}
\begin{document}
\pagestyle{headings}
\mainmatter
\def\ECCVSubNumber{1765}  

\title{Reducing Language Biases in Visual Question Answering with Visually-Grounded Question Encoder} 

\titlerunning{Visually Grounded Question Encoder for VQA}
%
\author{Gouthaman KV  \and 
Anurag Mittal}
%
%
\institute{Indian Institute of Technology Madras, India \\
\email{\{gkv,amittal\}@cse.iitm.ac.in}}
\maketitle

\begin{abstract}
Recent studies have shown that current VQA models are heavily biased on the language priors in the train set to answer the question, irrespective of the image. E.g., overwhelmingly answer ``what sport is" as ``tennis" or ``what color banana" as ``yellow." This behavior restricts them from real-world application scenarios. In this work, we propose a novel model-agnostic question encoder, Visually-Grounded Question Encoder (VGQE), for VQA that reduces this effect. VGQE utilizes both visual and language modalities equally while encoding the question. Hence the question representation itself gets sufficient visual-grounding, and thus reduces the dependency of the model on the language priors.  We demonstrate the effect of VGQE on three recent VQA models and achieve state-of-the-art results on the bias-sensitive split of the VQAv2 dataset; VQA-CPv2. Further, unlike the existing bias-reduction techniques, on the standard VQAv2 benchmark, our approach does not drop the accuracy; instead, it improves the performance. 
\end{abstract}

\section{Introduction}
Visual Question Answering (VQA) is a good benchmark for context-specific reasoning and scene understanding that requires the combined skill of Computer Vision (CV) and Natural Language Processing (NLP). Given an image and a question in natural language, the task is to answer the question by understanding cues from both the question and the image. Tackling the VQA problem requires a variety of scene understanding capabilities such as object and activity recognition, enumerating objects, knowledge-based reasoning, fine-grained recognition, and common sense reasoning, etc. Thus, such a multi-domain problem yields a good measure of whether computers are reaching capabilities similar to humans. 

With the success of deep-learning in CV and NLP, many datasets~\cite{vqa1,vqa2,tdiuc,vizwiz,gqa,zellers2019recognition}, and models~\cite{bottomup,mutan,murel,block,ban} have been proposed to solve VQA. Most of these models perform well in the existing benchmark datasets where the train and test sets have similar answer distribution. However, recent studies show that these models often rely on the statistical correlations between the question patterns  \& the most frequent answer in the train set, and shows poor generalized performance (i.e., poor performance on a test set with different answer distributions than the train set~\cite{gvqa}).
In other words, they are heavily biased on the language modality. E.g., most of the existing models overwhelmingly answer ``tennis" by seeing the question pattern ``what sport is.. " or answer the question ``what color of the banana?"  as ``yellow" even though it is a ``green banana," By definition and design, the VQA models need to merge the visual and textual modalities to answer the question. However, in practice, they often answer the question without considering the image modality, i.e., they tend to have less image grounding and rely more on the question. This undesirable behavior restricts the existing VQA models being applicable in practical scenarios. 

One reason for this behavior is the strong language biases that exist in the available VQA datasets~\cite{vqa1,vqa2}. Hence, the models trained on these datasets will often rely on such biases~\cite{gvqa,overcoming,rubi,hint,self_critical_bias,bias_aaai}. E.g., in the most popular and large VQAv2 dataset~\cite{vqa2}, a majority of the ``what sport is" question is linked to ``tennis." Hence, upon training, the model will blindly learn the superficial correlations between the ``what sport is" language pattern from the question, and the most frequent answer ``tennis." Unfortunately, it is hard to avoid these biases from the train set when collecting real-world samples due to the annotation cost. Instead, we require methods that force the model to look at the image and question equally to predict the answer. Since the main source of bias is the over-dependency of the models on the language side, some approaches such as~\cite{overcoming,rubi,bias_aaai}, tried to overcome this by reducing the contribution from the language side. On the other hand, some approaches~\cite{hint,gvqa,self_critical_bias} tried to improve the visual-grounding of the model in order to reduce the language-bias. All of these existing bias reduction methods cause a performance reduction in the standard benchmark VQA datasets.  

A common pipeline in existing VQA models is to encode the image and question separately and then fuse them to predict the answer. Typically the image is encoded using a pre-trained CNN (e.g., ResNet~\cite{resnet}, VGGNet~\cite{vgg}, etc. ), and the question is encoded by sending the word-level features to an RNN (GRU~\cite{gru} or LSTM~\cite{lstm}). In this approach, while encoding the question, it only considers the language modality, and such a question representation has not contained any distinguishable power based on the content in the image. In other words, the question representation is not grounded in the image. E.g., with this scheme, the question ``what sport is this?" has the same encoded representation irrespective of whether the image contains ``tennis," ``baseball," or ``skateboarding," etc. Since the majority of such questions are linked to the answer ``tennis" in the train set, the model will learn a strong correlation between the question pattern ``what sport is" to the answer ``tennis," irrespective of the image. This leads to overfitting the question representation to the most frequent answer, and the model will ignore the image altogether.  

\begin{figure}[t!]
\centering
\subfloat[A generic VQA model with a traditional question encoder.]{\includegraphics[width=0.45\columnwidth]{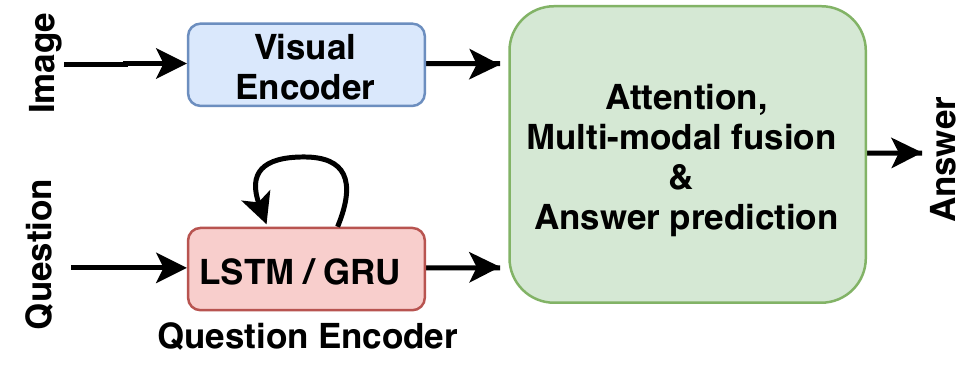}}\qquad
\subfloat[A generic VQA model with VGQE.]{\includegraphics[width=0.4\columnwidth]{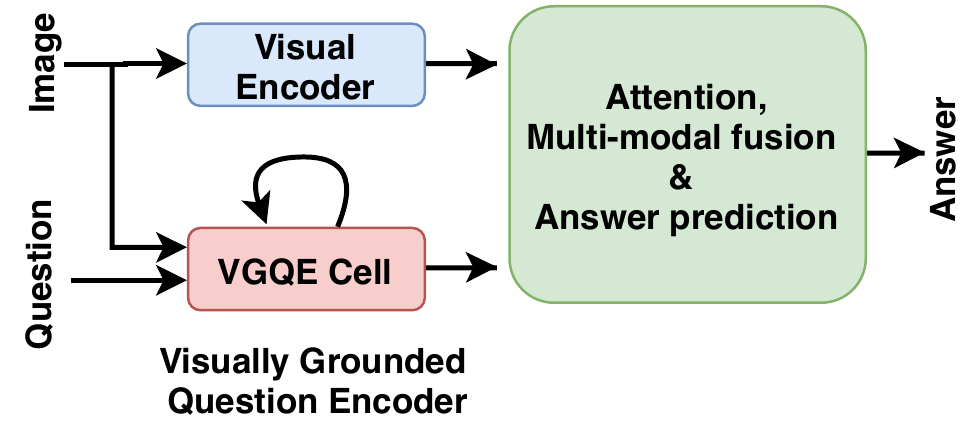}}
\caption{A VQA model with a traditional question encoder vs. with VGQE: A traditional question encoder uses only the language modality while encoding the question, whereas VGQE uses both language and visual modalities to encode the question.}
\label{fig:comparison}
\vspace{-0.3cm}
\end{figure}
In this paper, we propose a generic question encoder for VQA, the Visually-Grounded Question Encoder (VGQE), that reduces this problem. VGQE encodes the question by considering not only the linguistic information from the question but also the visual information from the image. A visual comparison of a generic VQA model with a traditional question encoder and with VGQE is shown in Fig~\ref{fig:comparison}. The VGQE explicitly forces the model to look at the image while encoding the question. For each question word, VGQE finds the important visual feature from the image and generates a visually-grounded word-embedding vector. This word embedding vector contains the language information from the question word and the corresponding visual information from the image. 
These visually-grounded word-embedding vectors are then passed to a sequence of RNN cells (inside VGQE) to encode the question. 
Since VGQE considers both the modalities equally, the encoded question representation will get sufficient distinguishing power based on the visual counterpart. For e.g., with VGQE, in the case of the ``what sport is this" question, the model can easily distinguish the question for ``baseball," ``tennis," or  ``skateboarding," etc. As a result, the question representation itself is heavily influenced by the image, and the learning of the correlation between specific language patterns and the most frequent answers in the train set can be reduced.

The VGQE is generic and easily adaptable to any existing VQA models, i.e., replace the existing traditional language-based question encoder in the model with VGQE.  In this paper, we demonstrate the ability of VGQE to reduce the language bias on three recent best performing baseline VQA models, i.e., MUREL~\cite{murel}, UpDn~\cite{bottomup} and BAN~\cite{ban}. We did extensive experiments on the VQA-CPv2 dataset~\cite{gvqa} and demonstrate the ability of VGQE to push the baseline models to achieve state-of-the-art results. The VQA-CPv2 dataset is specifically designed to test the VQA model's capacity to overcome the language biases from the train set. Further, we show the effect of VGQE in the standard VQAv2 benchmark as well. Unlike existing bias reduction techniques~\cite{gvqa,overcoming,rubi,bias_aaai,self_critical_bias}, VGQE does not show any drop in the accuracy on the standard VQAv2 dataset; instead, it improves the performance.

\section{Related Works}
A major role in the success of deep-learning models lies in the availability of large datasets. Most of the existing real-world datasets for various problems will have some form of biases, and it is hard to avoid this at the time of dataset collection, due to the annotation cost. Consequently, the models trained on these datasets often over-fit to the biases between the inputs and ground truth annotations from the train set. Researchers tried different procedures to mitigate this problem in various domains. For instance, some methods focused on the biases in captioning models~\cite{cap_bias}, gender biases in multi-label object classification~\cite{gender_bias}, biases in ConvNets and ImageNet~\cite{convnet_bias}, etc. Since our work focuses on reducing the language biases in VQA models, in the following, we discuss the related works lying on the same line. 

 \textbf{Approaches in the dataset side:} 
In VQA, various datasets have been proposed so far such as~\cite{vqa1,vqa2,tdiuc,genome}.  The main source of bias in these datasets are from the language side, i.e., the question. 
There exist strong correlations between some of the word patterns in the question and the frequent answer words in the dataset. For e.g., in the large and popular VQAv1 dataset~\cite{vqa1}, there exist strong language priors such as ``what color.." to ``white," ``is there.." to ``yes," ``how many.." to ``2", etc. To reduce the language biases in VQAv1, in~\cite{vqa2}, the authors introduced a more balanced VQAv2 dataset, by adding at least two similar images with different answers for each question. However, with this additional refinement also, there still exists a considerable amount of language biases that can be leveraged by the model during training. In~\cite{gvqa,explicit_bias,being_negative}, the authors empirically show that a question-only model (without using the image) trained on VQAv2 shows reasonable performance, indicating the strong language priors in the dataset. 
In~\cite{gvqa}, the authors show that since the train and test answer distributions of VQAv2 are still similar, and the models that solely memorize language biases in the train set show acceptable performance while testing. In this regard, they introduced a diagnostic dataset, the VQA-CP (VQA under Changing Priors), to measure the language bias learned by the VQA models.  This dataset is constructed with vastly different answer distributions
between the train and test splits. Hence the models that rely heavily on the language priors in the train set will show poor performance while testing. \cite{gvqa} empirically shows that most of the existing best performing VQA models on the VQAv2 dataset show a significant drop in the accuracy when tested on the new test split from VQA-CPv2.  E.g., a model with $\approx 66 \%$ accuracy in the standard VQAv2 shows only $\approx 40 \%$ accuracy in the bias sensitive VQA-CPv2. 

\textbf{Approaches in the architecture side:} 
Several approaches have been considered in the literature to remove such biases. In~\cite{gvqa}, the authors propose a specific VQA model built upon~\cite{san}, the GVQA model, consists of specially designed architectural restrictions to prevent the model from relying on the question-answer correlations.  This model is complicated in design and requires multi-stage training, which makes it challenging to adapt to existing VQA models. In~\cite{overcoming,rubi}, the authors propose different regularization techniques that can be applied via the question encoder to reduce the bias.
In~\cite{overcoming}, the authors added an adversary question-only branch to the model, and it is supposed to find the amount of bias that exists in the question. Then, a gradient negation of the question-only branch loss is applied to remove answer discriminative features from the question representation of the original model. They also propose a difference of entropy (DoE) loss captured between the output distributions of the VQA model and the additional question-only branch. In~\cite{rubi}, the authors propose a similar approach as in~\cite{overcoming}, named RUBi, where instead of using the DoE and question-only loss, they mask (using the \textit{sigmoid} operation) the output from the question-only branch and element-wise multiplying it with the output of the original model, to dynamically adapt the value of the classification loss to reduce the bias. However, both of these regularization approaches show a decrease in the performance on the standard VQAv2 benchmark, while reducing the bias. 
In~\cite{hint}, the authors propose a tuning method, called HINT, that is used to improve the visual-grounding of the existing model. Specifically, they tuned the model with manually annotated attention maps from the VQA-HAT dataset~\cite{vqa-hat}. Similar to this, in~\cite{self_critical_bias}, the authors propose a tuning approach called SCR, which uses additional manually annotated attention maps (from the VQA-HAT~\cite{vqa-hat}) or textual explanations (from the VQA-X dataset~\cite{vqa-x}) to improve the visual-grounding of the model. However, both of these approaches~\cite{hint,self_critical_bias} require additional manually annotated data, and collecting the same is expensive. Also, these approaches reduce the performance on the standard VQAv2. Recently, in~\cite{bias_aaai}, the authors proposed an approach based on language-attention and is built upon the GVQA model~\cite{gvqa}. The language attention module splits the question into three language phrases: question type, referring objects, and specific features of the referring object. These phrases further utilized to infer the answers from the image. They claim that splitting the question into different language phrases reduces the learning of the bias. However, this approach also shows a reduction in the performance on VQAv2.

The prior works are trying to overcome the bias either by reducing the contribution from the language side as in~\cite{overcoming,rubi,bias_aaai}, or by improving the visual-grounding by tuning with additional manually annotated data as in~\cite{hint,self_critical_bias}. One common drawback of all of these methods is that they show a significant drop in the performance on the standard VQAv2 benchmark while reducing the bias. On the contrary, our approach improves the representation power of the question encoder to make visually-grounded question encodings to reduce the bias.  The advantage of such an approach is that it is not only model-agnostic but also does not sacrifice the performance on the standard VQAv2 benchmark.
Also, it does not require any additional manually annotated data or tuning.

 \textbf{Traditional question encoders and pitfall:}
 The widely adopted question encoding scheme in VQA is passing the word-level features to a recurrent sequence model (LSTM~\cite{lstm} or GRU~\cite{gru}) and taking the final state vector as the encoded question. Some early works use one-hot representations of question words and pass through an LSTM such as in ~\cite{vqa1,mcb,mfb}, or GRU as in~\cite{mutan}. The drawback of using the one-hot vector representations of question words is the manual creation of the word vocabulary.  Later, the usage of pre-trained word embedding vectors such as GloVe~\cite{glove}, BERT~\cite{bert}, etc., becomes popular, since they do not require any word vocabulary and also rich in linguistic understanding. This question encoding scheme, i.e., pass the pre-trained word embeddings of the question words to some RNNs, is the currently popular approach in VQA~\cite{bottomup,ban,murel,block,improved,mlin}. However, recently, with the success of Transformer based models in various NLP tasks~\cite{bert}, usage of Multi-modal Transformers (MMT) are becoming popular, such as ViLBERT~\cite{vilbert}, LXMERT~\cite{lxmert}, etc. These are co-attention models working on top of Transformers~\cite{transformer}, where while encoding the question, the words are prioritized with the guidance of the visual information.

The above-mentioned question-encoding schemes only use the language modality, i.e., the word-embeddings of the question words. As a result, the encoded question contains only the linguistic information from the question, and it cannot distinguish the questions based on their visual counterpart. This situation forces the model to learn the unwanted correlations between the language patterns in the questions, and the most frequent answers in the train set without considering the image which leads to a language side biased model. On the contrary, our question encoder, the VGQE, considers both visual and language modalities equally and generate visually-grounded question representations. A VQA model with such a question representation can reduce the over-dependency on the language priors in the train set. 

The VGQE is also related to approaches such as FiLM~\cite{film}, in terms of the early usage of the complementary information. The FiLM uses the question context to influence the image encoder, whereas VGQE uses the visual information inside the question encoder.

\section{Visually-Grounded Question Encoder (VGQE)}
We follow the widely-adopted RNN based question encoding scheme in VQA as the base.
In VGQE, instead of the traditional RNN cell, a specially designed cell called the VGQE cell is used. A VGQE cell takes the word embedding of the current question word and finds its relevant visual counterpart feature from the image, then creates a visually-grounded question word embedding. This new word embedding is passed to an RNN cell to encode the sequence information.  

Before going into the VGQE cell details, we explain the image and question feature representations used by the model. The image is represented by two sets of features $V=\{v_i\in \mathbb R^{d_v}\}_{i\in[1,k]}$ and  $L=\{l_i\in \mathbb R^{d_w}\}_{i\in[1,k]}$ corresponding to the CNN features of objects and embeddings of the class labels respectively, where $k$ is the total number of objects. The question is represented as a sequence of words with corresponding word embedding vectors, the word embedding of the $t^{th}$ question word is being denoted by $q_t\in \mathbb R^{d_w}$. 
\begin{figure}[t!]
    \centering
    \includegraphics[height=7cm]{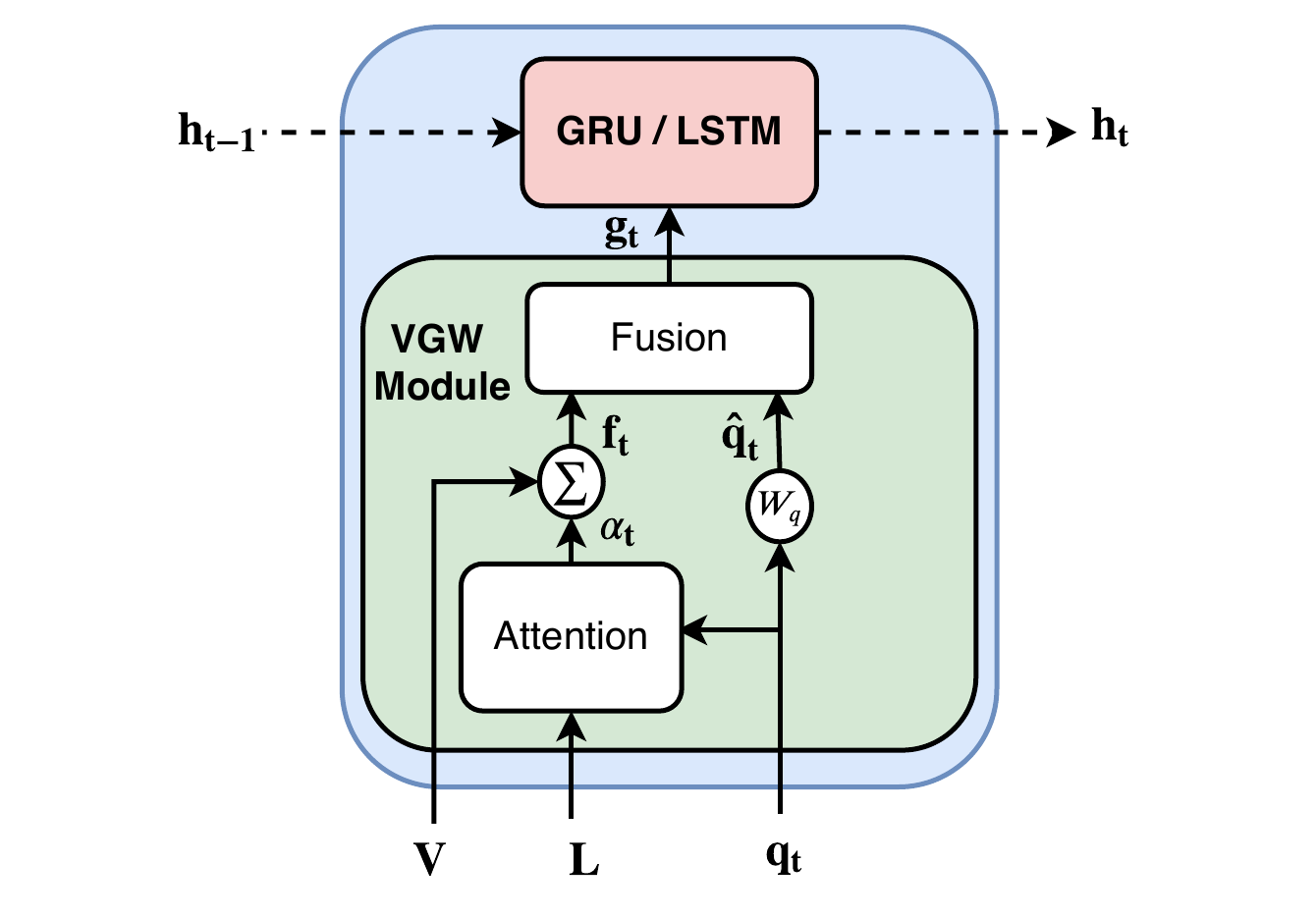}
    \caption{ VGQE cell at time $t$:  VGW module finds the visually-grounded word embedding $g_t$ for the current question word $q_t$ and the RNN cell (GRU or LSTM) encodes the sequential information in the question. $V$ and $L$ are the sets of object-level features.}
    \label{fig:vgqe}
\end{figure}

\subsection{VGQE cell}
The basic building block of the proposed question encoder is the VGQE cell. This is analogous to the RNN cell (LSTM~\cite{lstm} or GRU~\cite{gru}) in the traditional approach. An illustration of the VGQE cell at time $t$ is shown in Fig.~\ref{fig:vgqe}. Each VGQE cell takes the object-level features $V$, $L$, the current question word embedding $q_t$ and the previous state context vector $h_{t-1}$, and then outputs the current state context vector $h_t$. The context vector $h_t$ is then pass to the next VGQE cell in sequence. Mathematically we represent a VGQE cell as follows:
\begin{equation}
    h_t=\text{VGQE} (V,L,q_t,h_{t-1})
\end{equation}
A VGQE cell consists of two modules: the Visually-Grounded Word embedding (VGW) module and a traditional RNN cell (GRU~\cite{gru} or LSTM~\cite{lstm}). The VGW module is responsible for finding the visually-grounded word embedding vector $g_t$ for the current question word embedding $q_t$. Then, $g_t$ is passed to the RNN cell to encode the question sequence information.
\subsubsection{VGW module:} This module is the crux of the VGQE cell, where the visually-grounded word embedding $g_t$ for the current question word $q_t$ is extracted. It works  in two stages: 1)\textit{Attention} and 2)\textit{Fusion}. The first stage calculates the visual counterpart feature $f_t$ of the question word embedding $q_t$, while in the fusion stage, $f_t$ and $q_t$ are fused to generate the visually-grounded word embedding vector $g_t$.

\noindent\textbf{Attention:} This module takes the set of object-label features $L$ and the current question word $q_t$ as inputs and outputs a relevance score vector $\alpha_t \in \mathbb R^k$. Each of the $k$ values in $\alpha_t$ tells the relevance of the corresponding object from the image in the context of the current question word $q_t$.  Then, one single visual feature vector $f_t$ is extracted by taking the weighted sum over the set of object CNN features $V$, where the weights are defined by $\alpha_t$. Mathematically, the above steps are formulated as follows: 

\begin{align}\label{eq:att}
   f_t&=\sum_{i=1}^k \alpha_t[i]*v_i \\ 
   \nonumber
    \alpha_t&= softmax(w_a^T (W(\mathbf{L}*(\mathds{1} q_t))^T)) 
\end{align}
where $*$ is element-wise multiplication, $v_i \in V$ is the $i^{th}$ object CNN feature vector,  $\alpha_t \in \mathbb R^k$ and  $f_t\in \mathbb R^{d_v}$ are the object-relevance score vector and the visual counter part feature for the current question word $q_t$ respectively, $\alpha_t[i]$ is the relevance of the $i^{th}$ object at time $t$, $\mathbf{L} \in \mathbb R^{k \times d_w}$ is the matrix representation of the set of object-label features, $w_a\in \mathbb R^{d_w}$ and $W\in \mathbb R^{d_w \times d_w}$ are learnable parameters and $\mathds{1}$ is a column vector of length $k$ consists all ones.

\noindent\textbf{Fusion:} In this stage, the word-level visual feature $f_t$ is grounded to the word-level language feature $q_t$. This results in the visually-grounded word embedding vector $g_t$. 
For fusion, any multi-modal fusion function $F_m$ such as element-wise multiplication, bi-linear fusion~\cite{mcb,mutan,mlb,mfb,block} etc. can be used. In this paper, we use the BLOCK fusion as $F_m$, which is a recently proposed best performing bi-linear multi-modal fusion method~\cite{block}. The formal representation of the visually-grounded word embedding vector $g_t$ is as follows: 
\begin{equation}
    g_t=F_m(f_t,\hat q_t;\Theta) 
\end{equation}
where $F_m$ is the multi-modal fusion function (in our case BLOCK) with $\Theta$ as the learnable parameters, the vector $\hat q_t= W_q(q_t)$  is the fine-tuned word embedding vector with $W_q$ as the learnable parameters of a two-layer network that projects from $d_w$-space to $d$-space, and  $f_t$ is the visual feature as defined in Eq. \eqref{eq:att}. The fused vector $g_t$ contains relevant language information from the word embedding $q_t$ and visual information from $f_t$; in other words, $g_t$ is a visually-grounded word embedding vector. For the same question word (e.g., banana), the $g_t$ vector will vary according to their visual counterpart (as a result, the ``green banana" and ``yellow banana" will get distinguishable representations in the question.). 

The output from the VGW module, the visually-grounded word embedding vector $g_t$ along with the previous state context vector $h_{t-1}$, is then passed to a traditional RNN cell to generate the current state question context vector $h_t$. 

\subsection{Using VGQE cell to encode the question}
\begin{figure}[t]
    \centering
    \includegraphics[width=0.9\columnwidth]{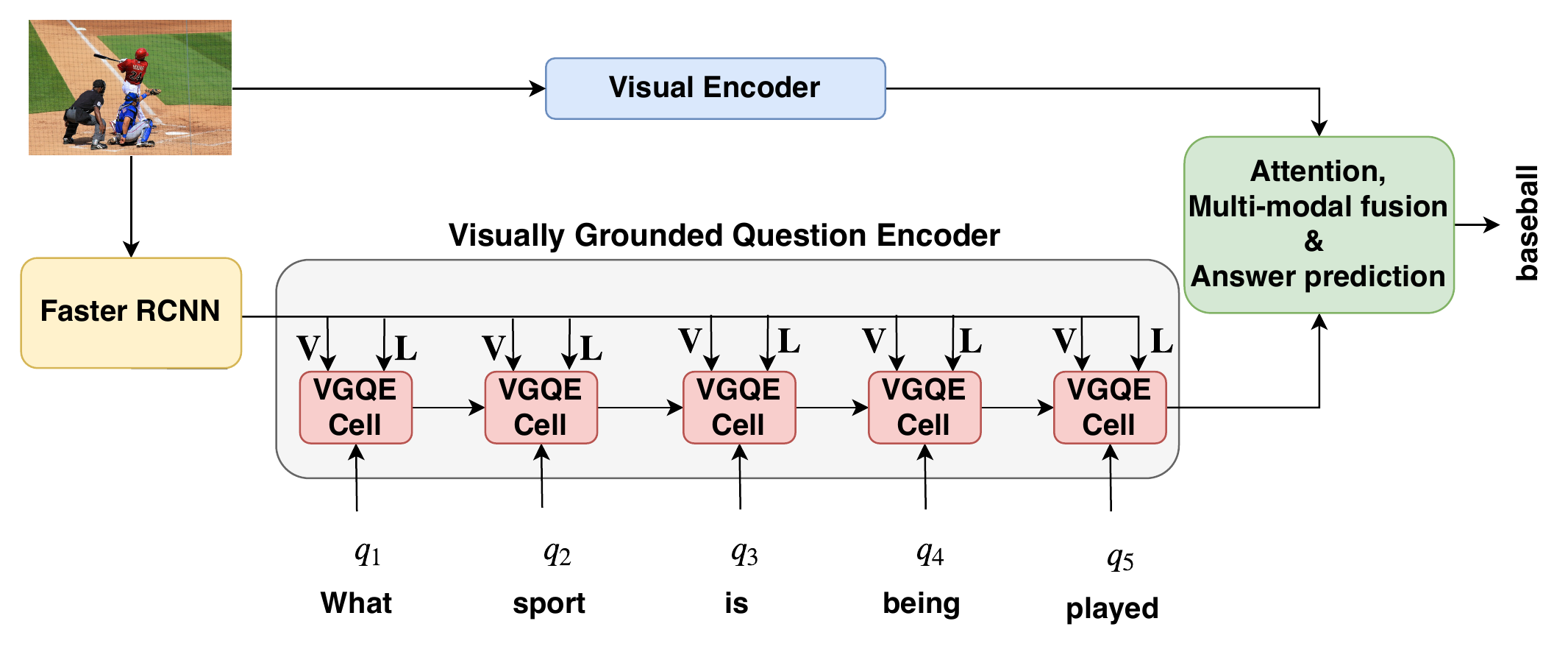}
    \caption{An illustration of a generic VQA model with VGQE}
    \label{fig:generic_model}
\end{figure}

In the proposed question encoder, we use the VGQE cell instead of the traditional RNN cell. An illustration of a generic VQA model with VGQE is shown in Fig.~\ref{fig:generic_model}.  The visual encoder encodes the image as in the original model. The question is encoded using VGQE instead of the existing question encoder. The question word embeddings are passed through a sequence of VGQE cells along with the object-level features ($V$ and $L$) and take the final state representation as the encoded question. In this paper, we use GRU as the RNN cell inside the VGQE cell. Other question-encoder adaptations using variants of RNN, such as LSTM, are also possible, with appropriate changes in the RNN cell of VGQE. Since, in VGQE, the question words are grounded on its visual counterpart, the encoded questions itself contain sufficient visual-grounding. Hence, they become robust enough to reduce the correlation between specific language patterns and the most frequent answers in the train set.

\subsection{Baseline VQA architecture}
In this paper, we adopt the same baseline as used in prior work~\cite{rubi}, to perform various experiments, also to make a fair comparison.  This baseline is a simplified version of the MUREL VQA model~\cite{murel}. The word embeddings are passed to a GRU~\cite{gru} to encode the question. Then, each of the object CNN features $v_i \in V$ are bi-linearly fused (using BLOCK fusion~\cite{block}) with the encoded question to get the question-aware object features. A max-pool operation over these features gives a single vector that later used by the answer prediction network.

\section{Experiments and Results}
\subsection{Experimental setup}
We train and evaluate the models on the VQA-CPv2 dataset~\cite{gvqa}. This dataset is designed to test the robustness of VQA models in dealing with language biases. The train and test sets of VQA-CPv2 have totally different answer distributions; hence a model that strongly depends on the language priors in the train set will perform poorly on the test set. We also use the VQAv2 dataset to train and evaluate the models on the standard VQA benchmark. We use the VQA accuracy~\cite{vqa1} as the evaluation metric.

We use the object-level CNN features provided in~\cite{bottomup} as $V$ ($d_v=2048$). 
We use the pre-trained Glove~\cite{glove} for the object-label features $L$ and question word embeddings $q_t$ ($d_w=300$). 
We train the model using Adamw~\cite{adamw} with $weight decay = 2*10^{-5}$ and cross-entropy loss. 
Further implementation details are provided in the supplementary material.

\begin{table}[t!]
\begin{floatrow}
\ttabbox{
    \begin{tabular}{lcccc}
    \toprule
    Model & Overall& Yes/No & Number& Other \\
    \midrule
    GVQA~\cite{gvqa}&31.30&57.99&13.68&22.14\\
         RAMEN~\cite{ramen}&39.21&-&-&-\\
         BAN~\cite{ban} & 39.31&-&-&-\\
         MUREL~\cite{murel} & 39.54&42.85&13.17&45.04\\
         UpDn~\cite{bottomup}& 39.74& 42.27&11.93&46.05 \\
         UpDn+Q-Adv+DoE~\cite{overcoming} & 41.17 & 65.49 &15.48 & 35.48\\ 
         UpDn+HINT~\cite{hint}&46.73&67.27&10.61&45.88\\
         UpDn+LangAtt~\cite{bias_aaai}&48.87&70.99&18.72&45.57\\
         UpDn+SCR (VQA-X)~\cite{self_critical_bias}&49.45&\textbf{72.36}&10.93&\textbf{48.02}\\
         Baseline~\cite{rubi} &38.46& 42.85&12.81&43.20 \\ 
         Baseline+RUBi~\cite{rubi} &47.11& 68.65 & 20.28&43.18 \\
         \midrule
         Baseline+VGQE& \textbf{50.11} & 66.35 & \textbf{27.08} & 46.77 \\
         \bottomrule
    \end{tabular}}
    {\caption{Comparison with existing models on VQA-CPv2.}} \label{tab:sota_comparison}
    \end{floatrow}
\end{table}

\subsection{Results}
\textbf{Comparison with state-of-the-art models:} In Table~\ref{tab:sota_comparison}, we compare VGQE against existing state-of-the-art bias reduction techniques in VQA-CPv2. We can see that our approach achieves a new state-of-the-art. With VGQE, we improved the baseline model (that uses the traditional GRU based question encoder) accuracy from $38.46$ to $50.11$. This performance also corresponds to a notable improvement from the RUBi approach~\cite{rubi}, where they also use the same baseline. Our approach outperforms the UpDn based approaches~\cite{overcoming,hint,bias_aaai,self_critical_bias} as well. Comparing with GVQA, a specific model designed for VQA-CP,  our generic approach outperforms with a $+18.81$ gain in the accuracy.

\begin{figure}[t]
\CenterFloatBoxes
\begin{floatrow}
\ttabbox
  {\begin{tabular}{l|c|c}
        \toprule
        Question Type& Baseline & +VGQE\\
        \midrule
        Do you & 30.7 & 93.19 \\
        Can you & 29.04 & 52.54 \\
        What are the& 39.8 & 48.52 \\
        What color are the & 28.32 &64.42 \\
        What sport is & 88.39 & 93.19 \\
        What is the person & 57.40&63.65\\
        What time & 39.28 & 57.88\\
        What room is & 80.34 & 92.87 \\
        What is in the & 29.34 & 35.84 \\
        Where are the & 19.87 & 31.88 \\
        What brand & 28.54 &47.19 \\
        How many & 12.91 & 19.65\\
        How many people &36.62&50.03\\
        Does the & 31.91 &77.39\\
        Which & 26.73 & 39.0 \\
        Why & 12.28 & 14.76 \\
        \bottomrule
    \end{tabular}
  }
  {\caption{Performance comparison between the baseline and VGQE corresponding to some of the question-types in the VQA-CPv2 test set. VGQE reduces the learning of unwanted question pattern-answer correlations and improves the performance. All numbers are VQA accuracy values. }\label{tab:qtype_comparison}}
 \killfloatstyle
\ffigbox
  {
  \centering
 \includegraphics[width=0.4\textwidth,height=8cm]{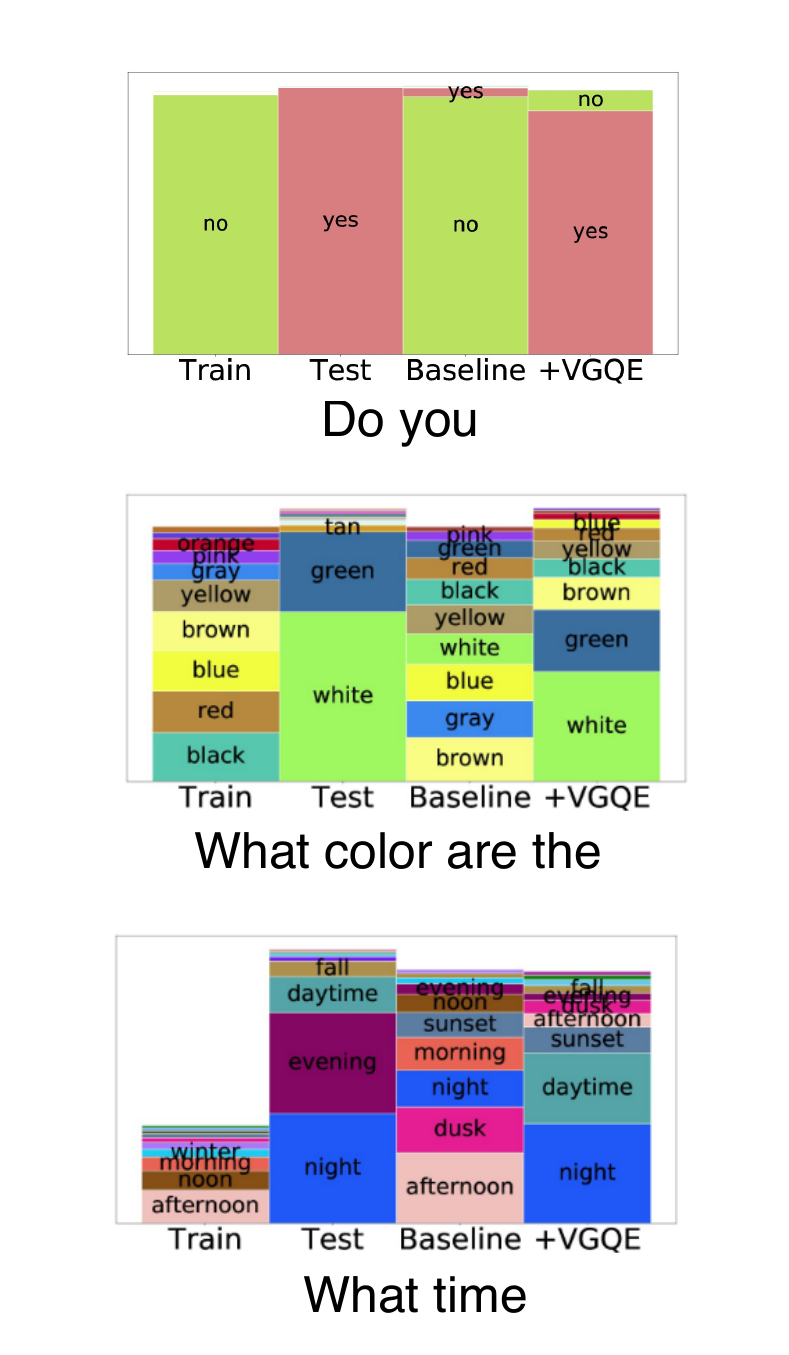}}
  {\caption{Answer distributions from train \& test sets of VQA-CPv2, and outputs of baseline \& baseline+VGQE models, for some question types from Table~\ref{tab:qtype_comparison}. We can see that VGQE helps the model to reduce the learning of bias from the train set.}\label{fig:ans_distribution}}
\end{floatrow}
\end{figure}

\textbf{Question-type-wise results:} In Table~\ref{tab:qtype_comparison}, we show the comparison of some of the question-type-wise results of the baseline and with VGQE (+VGQE) models.  The baseline model shows comparatively poor performance than the model with VGQE. Since most of the time, the baseline model with a traditional question encoder tries to memorize the language biases in the train set. We can see that, in all the question-types, the incorporation of VGQE reduces such biases and improves performance. To further clarify this, in Fig.~\ref{fig:ans_distribution}, we visualize the answer distributions of some of the question-types in the VQA-CPv2 train \& test sets and outputs of the baseline \& baseline+VGQE models (less frequent answers are ignored for better visualization). It is clear from the visualizations that the baseline model learns the language biases in the train set and predicts the answer without considering the image. The incorporation of VGQE helps to reduce such biases and to answer by grounding the question onto the image. E.g., from Fig.~\ref{fig:ans_distribution}, consider the case of the ``Do you" type questions. Most of such questions are linked to the answer ``no" in the train set; in other words, there is a strong correlation between the ``Do you" language pattern and the answer ``no." We can see that most of the time, the baseline model predicts the answer "no" upon seeing the pattern ``Do you," without considering the image. This indicates that the baseline model relies on the strong correlation between the language pattern ``Do you" and the answer ``no," rather than looking at the image. In contrast, the incorporation of VGQE helps to reduce such learning of biases and to predict the answer by considering the image as well. 

On comparing the performance gain among the question types, we can see that the questions that start with "why" show relatively less improvement. Since these questions usually require common sense reasoning (e.g., Why people wear the hat?), and in such cases, the VGQE cannot improve much from the baseline.

\begin{figure}[t!]
    \centering
    \setlength\tabcolsep{0.5pt}
     \begin{tabular}{ccc}
    \includegraphics[width=0.3\columnwidth,height=2cm]{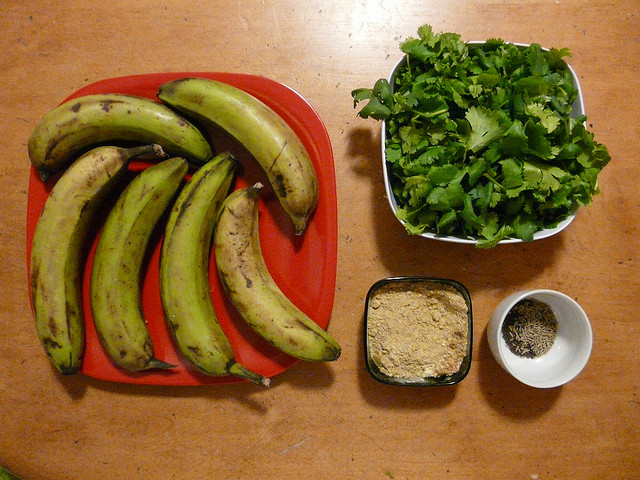}& \includegraphics[width=0.3\columnwidth,height=2cm]{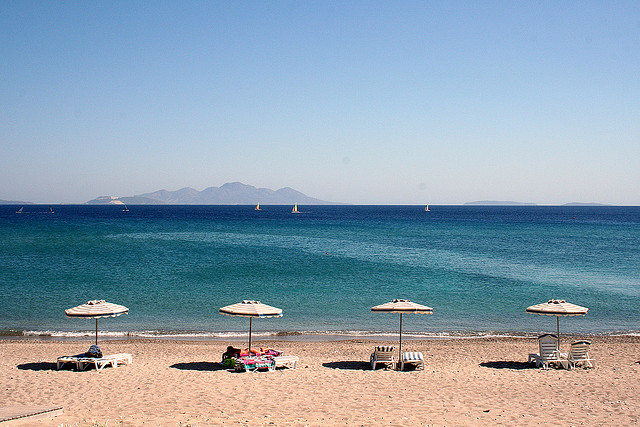}& \includegraphics[width=0.3\columnwidth,height=2cm]{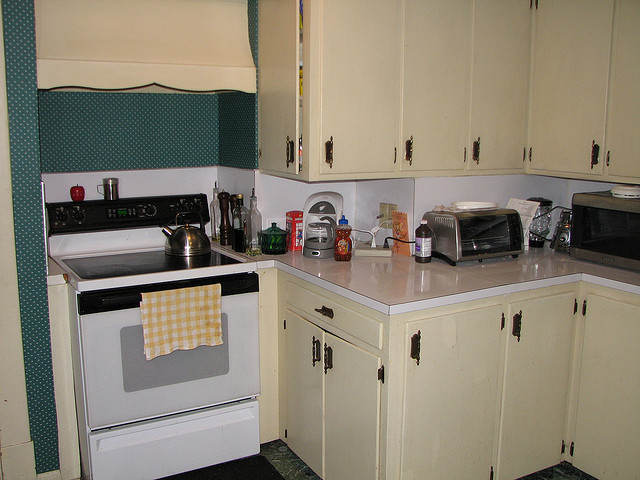}\\ 
   \makecell[c]{\textbf{Q:} What color are the \\ bananas. \textbf{GT:} green,\\ 
    \textbf{Ours:}green,\textbf{Base:} yellow.} & \makecell[c]{\textbf{Q:} What color are the \\ umbrellas. \textbf{GT:} white, \\
     \textbf{Ours:}white,\textbf{Base:} blue.}&\makecell[c]{\textbf{Q:} What color are the \\cabinets.
    \textbf{GT:} white, \\ \textbf{Ours:}white, \textbf{Base:} brown.}\\ \\
    \multicolumn{3}{c}{(a) \textbf{Question Pattern:} What color are the} \\ \\
        \includegraphics[width=0.3\columnwidth,height=2cm]{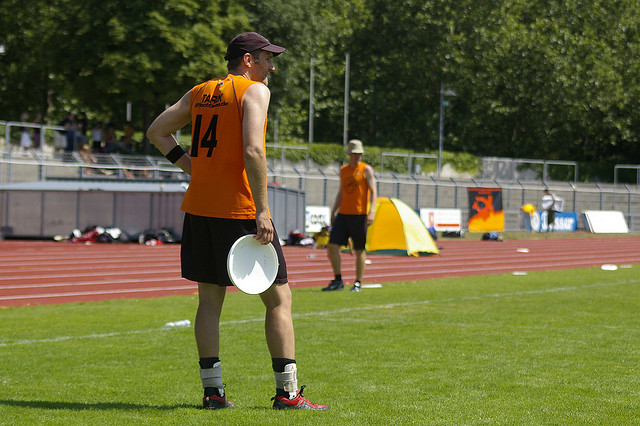}&\includegraphics[width=0.3\columnwidth,height=2cm]{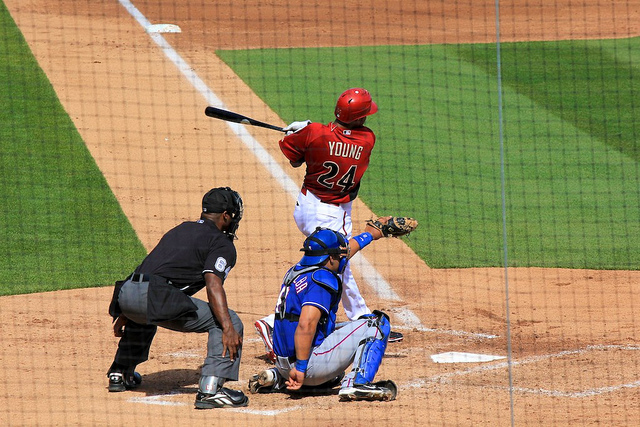}&\includegraphics[width=0.3\columnwidth,height=2cm]{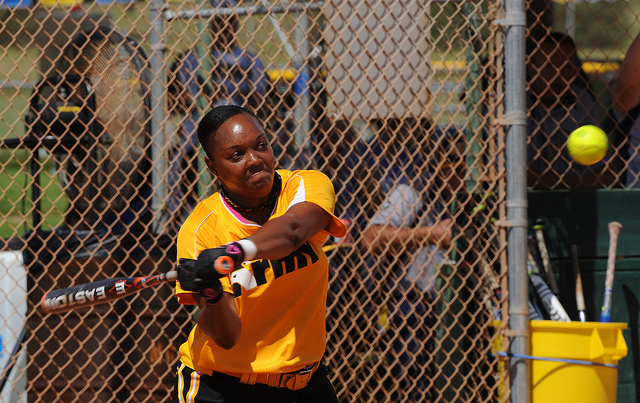}\\
    \makecell[c]{\textbf{GT:} frisbee, \\ \textbf{Ours:} frisbee, \\ \textbf{Base:} soccer} & \makecell[c]{\textbf{GT:} baseball, \\ \textbf{Ours:} baseball,\\ \textbf{Base:} tennis}&\makecell[c]{\textbf{GT:} baseball, \\ \textbf{Ours:} baseball,\\ \textbf{Base:} tennis}\\
    
    \multicolumn{3}{c}{(b) \textbf{Common Question:} What sport is being played?}\\ \\
    \includegraphics[width=0.3\columnwidth,height=2cm]{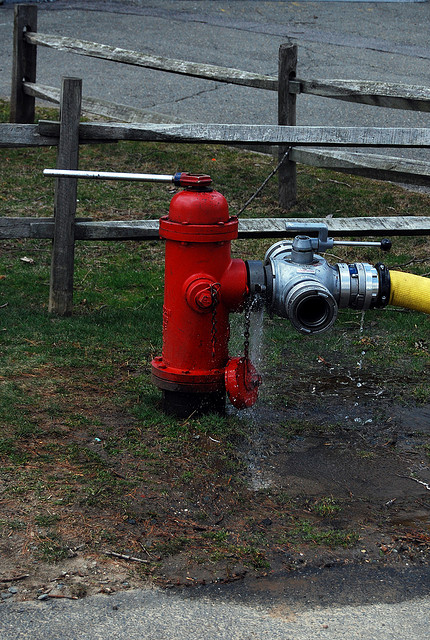}&\includegraphics[width=0.3\columnwidth,height=2cm]{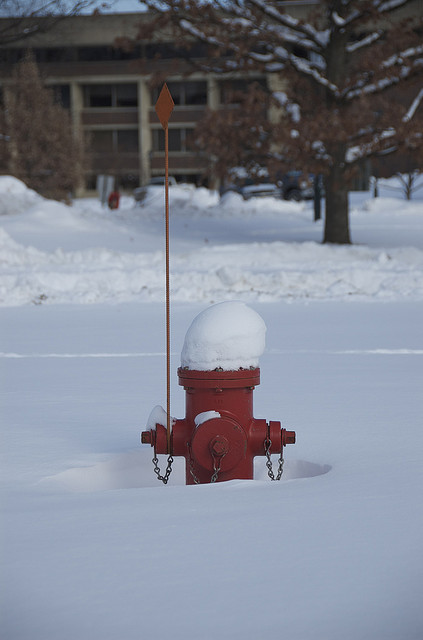}& \includegraphics[width=0.3\columnwidth,height=2cm]{images/COCO_val2014_000000123053.jpg} \\
    \makecell[c]{\textbf{GT:} red, \\ \textbf{Ours:} red, \textbf{Base:} yellow}&\makecell[c]{\textbf{GT:} red,\\ \textbf{Ours:} red, \textbf{Base:} white.}&\makecell[c]{\textbf{GT:} red,\\ \textbf{Ours:} red, \textbf{Base:} black}\\
    \multicolumn{3}{c}{(c) \textbf{Common Question:} What color is the fire hydrant?} \\
     \end{tabular}
    \caption{Some qualitative comparisons of our model (Baseline+VGQE) with Baseline. \textbf{Ours}, \textbf{Base} and \textbf{GT} represents the baseline+VGQE, baseline and the ground truth, respectively. Best viewed in color.}
    \label{fig:qual_results}
\end{figure}

\textbf{Qualitative results:}
In Fig.~\ref{fig:qual_results}, we show some qualitative comparisons between the baseline (Base) and baseline+VGQE (Ours) models. Fig~\ref{fig:qual_results} (a), shows the case of the ``What color are the" question pattern, where the train set has heavy biases towards the colors ``black", ``red",  ``blue",``brown", and ``yellow" (see Fig.\ref{fig:ans_distribution}). In Fig~\ref{fig:qual_results} (b) and (c), we show the cases of the same question asked for different images. The question ``What sport is being played?" has a bias on "tennis" ($\approx 63.2 \%$) and ``soccer" ($\approx 17.5 \%$) whereas ``What color is the fire hydrant?" is biased towards the colors ``yellow" ($\approx 46 \%$), ``white" ($\approx 12 \%$) and ``black" ($\approx 10 \%$) in the train set~\cite{gvqa}.  These visualizations further show that incorporating VGQE helps the baseline model to reduce the learning of biases.

\begin{figure}
    \setlength\tabcolsep{0.5pt}
    \begin{tabular}{cccccc}
     \includegraphics[width=0.15\columnwidth,height=2cm]{images/vqa_model_baseball.jpg} &
      \includegraphics[width=0.15\columnwidth,height=2cm]{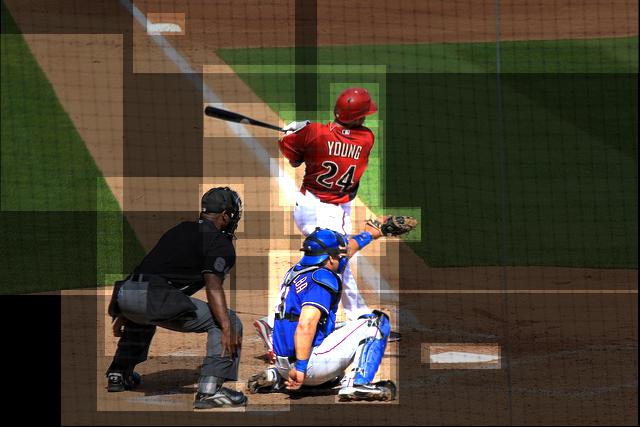} &  \includegraphics[width=0.15\columnwidth,height=2cm]{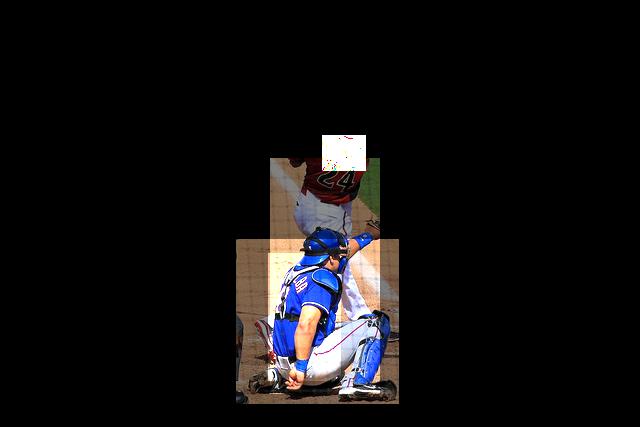}&\includegraphics[width=0.15\columnwidth,height=2cm]{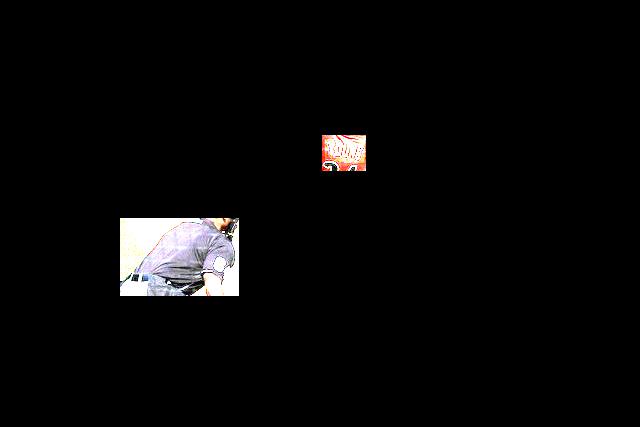}& \includegraphics[width=0.15\columnwidth,height=2cm]{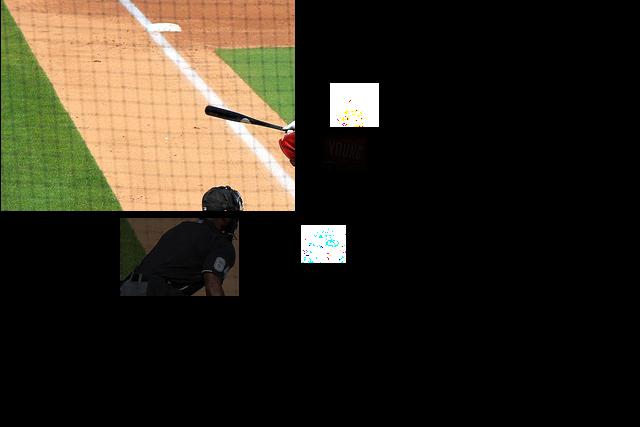}& \includegraphics[width=0.15\columnwidth,height=2cm]{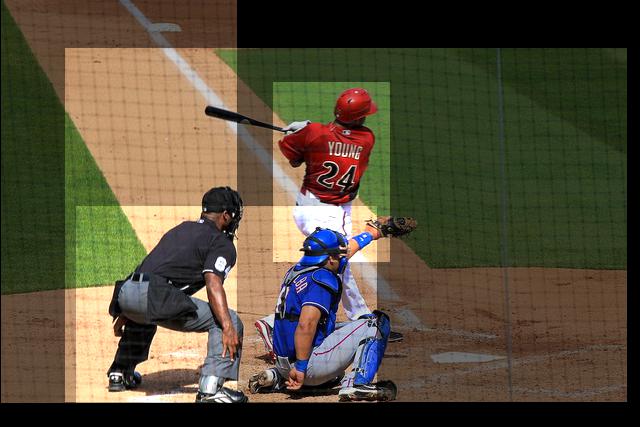}\\
      \makecell[c]{Input image} & \textbf{what}& \textbf{sport} &
      \textbf{is}&\textbf{being}&\textbf{played}\\
      
      \includegraphics[width=0.15\columnwidth,height=2cm]{images/COCO_train2014_000000023287.jpg} &
      \includegraphics[width=0.15\columnwidth,height=2cm]{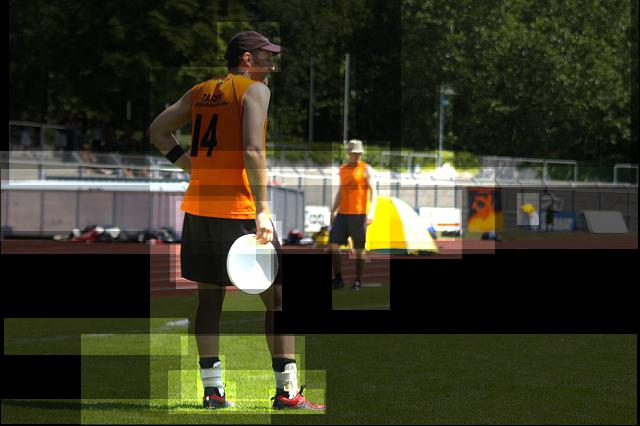} &  \includegraphics[width=0.15\columnwidth,height=2cm]{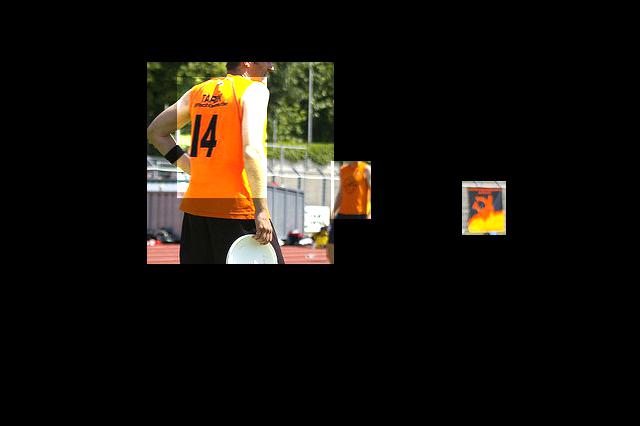}&\includegraphics[width=0.15\columnwidth,height=2cm]{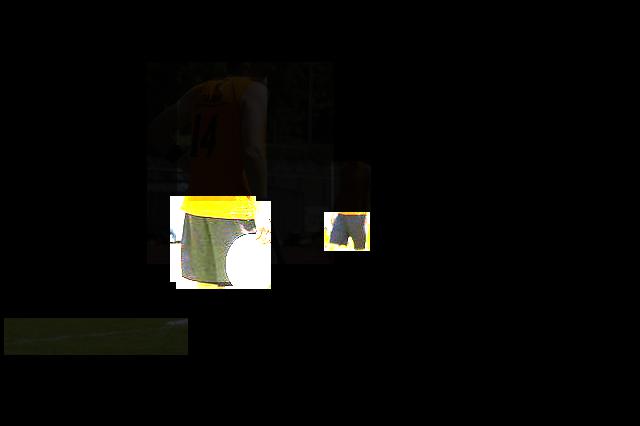}& \includegraphics[width=0.15\columnwidth,height=2cm]{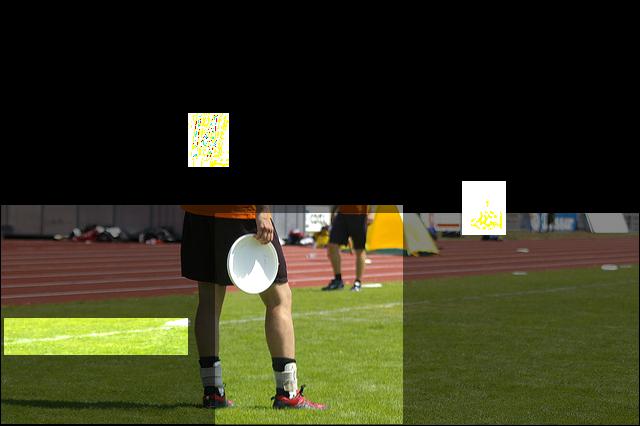}& \includegraphics[width=0.15\columnwidth,height=2cm]{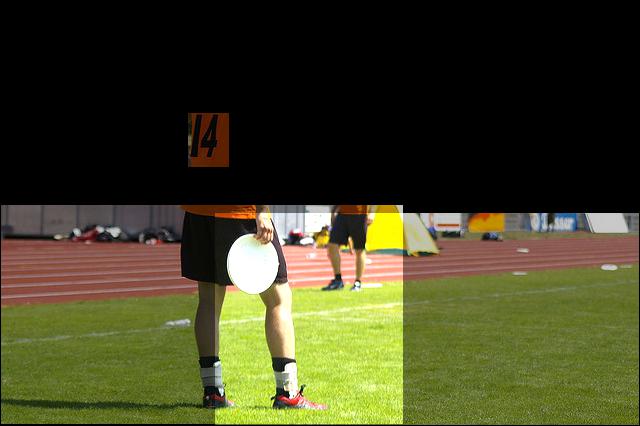}\\
      \makecell[c]{Input image} & \textbf{what}& \textbf{sport}&
      \textbf{is}&\textbf{being}&\textbf{played}\\
    \end{tabular}
    \caption{A visualization of the grounded visual regions at each time step of VGQE with the same question (``What sport is being played?") but different images. The same words will get different visually-grounded embeddings based on the visual counterpart.}
    \label{fig:vgqe_vis}
\end{figure}

In Fig.~\ref{fig:vgqe_vis}, we show a visualization of the grounded image regions at each time step of VGQE with the same question (``What sport is being played?") but different images. We can see that the question words are grounded onto the relevant visual counterparts. Hence, the same words will get different representations based on the image. For e.g., the word ``sport" is grounded on the "baseball" player in the first case, whereas it is grounded on the "frisbee" player in the second case. Hence, even though the question is the same, the VGQE generates different visually-grounded question representations.

\begin{table}[t]
    \CenterFloatBoxes
      \begin{floatrow}
\ttabbox
  { \begin{tabular}{lcc}
     \toprule
     Model & VQA-CPv2 & VQAv2 val\\ 
     \midrule
     UpDn~\cite{bottomup} & 39.74 & 63.48 \\
     UpDn+Q-Adv~\cite{overcoming} & 40.08 & 60.53\\
     UpDn+DoE~\cite{overcoming} & 40.43 & 63.43\\
     UpDn+Q-Adv+DoE~\cite{overcoming} & 41.17 & 62.75 \\
     UpDn+RUBi~\cite{rubi} & 44.23 & - \\
     UpDn+HINT~\cite{hint} &46.73&63.38\\
     UpDn+LangAtt~\cite{bias_aaai}&48.87&57.96\\
     UpDn+SCR~\cite{self_critical_bias}& 48.47&62.3\\
     UpDn+SCR (HAT)~\cite{self_critical_bias}& 49.17&62.2\\
         UpDn+SCR (VQA-X)~\cite{self_critical_bias}&\textbf{49.45}&62.2\\
     \midrule
     UpDn+VGQE & 48.75 &\textbf{64.04} \\
     \bottomrule
     BAN~\cite{ban,rubi} & 39.31 & 65.36 \\
     BAN+VGQE  & \textbf{50.00}& \textbf{65.73}\\
     \bottomrule
\end{tabular}} {\caption{Effect of VGQE on UpDn~\cite{bottomup} and BAN~\cite{ban} models, on VQA-CPv2 and VQAv2 val.} \label{tab:adaptation}} 

\ttabbox{\begin{tabular}{l|c}
    \toprule
         Model & VQAv2 val \\
         \midrule
     Baseline~\cite{rubi}&63.10 \\
        +RUBi~\cite{rubi}&61.16 \\
         \midrule
        +VGQE & \textbf{63.18}\\
         \bottomrule
    \end{tabular}}{\caption{Comparison of performance of the baseline model with RUBi and VGQE  on VQAv2 val set. RUBi shows a drop in the accuracy whereas VGQE does not show any drop.} \label{tab:rubi_vqav2}}
\end{floatrow}
\end{table}
\subsection{Performance of VGQE on other baselines} \label{exp:model_agnostic}
The proposed question encoder, the VGQE, can be easily incorporated into any existing VQA model, i.e., replace the existing question encoder with VGQE (See Fig.~\ref{fig:generic_model} for an illustration). We have already shown the effectiveness of VGQE on the MUREL baseline (see Table~\ref{tab:sota_comparison}). In this section, we show (see Table~\ref{tab:adaptation}) the impact of VGQE on two more recent best performing models as well:
\begin{itemize}
    \item \textbf{UpDn~\cite{bottomup}:} This VQA model is based on question guided visual attention. The word embeddings are pass to a GRU to encode the question. The image is encoded using visual attention guided by the question representation, over the set of object CNN features $V$. Then, both encoded image and question are combined using element-wise multiplication and pass to the answer predictor. 
    \item \textbf{BAN~\cite{ban}:} This model is based on bi-linear co-attention maps between the objects and question words. The question word embeddings are passed through a GRU. Then all the GRU cell outputs and the object-level CNN features are given to a co-attention module, the BAN. This module will give a combined vector which later used by the answer predictor.
\end{itemize}

In both models, we replaced the traditional GRU-based question encoder with VGQE. 
Note that, unlike other model-agnostic bias reduction methods, VGQE does not require a new question-only branch as in~\cite{overcoming,rubi} or tuning with additional annotated data as in~\cite{hint,self_critical_bias}, while training. The results are shown in Table~\ref{tab:adaptation}. We can see that, in both models, VGQE consistently improves the performance ($39.74$ to $48.75$ in UpDn~\cite{bottomup} and $39.31$ to $50.0$ in BAN~\cite{ban}) on VQA-CPv2, showing that here also VGQE reduces the language-bias.

\subsection{Performance of VGQE on the standard VQAv2 benchmark}
Existing bias reduction techniques such as~\cite{overcoming,rubi,bias_aaai,hint,self_critical_bias} show a reduction in the performance on the standard VQAv2 benchmark.
 For instance, in Table~\ref{tab:adaptation}, the best performing UpDn based models on VQA-CPv2, such as UpDn+LangAtt~\cite{bias_aaai} reduce the accuracy in the VQAv2 val set from $63.48$ to $57.96$ \& UpDn+SCR~\cite{self_critical_bias} reduces it from $63.48$ to $62.2$. Note that, UpDn+SCR reduces the performance even though it uses additional manually annotated data (HAT or VQA-X).
 Also, in Table~\ref{tab:rubi_vqav2}, adding RUBi to the baseline model decreases the accuracy from $63.10$ to $61.16$. On the contrary, VGQE does not sacrifice the performance on the standard VQAv2 benchmark. Interestingly, VGQE slightly improves the performance of the respective models; from $63.48$ to $64.04$, $63.10$ to $63.18$ and $65.36$ to $65.73$ in UpDn~\cite{bottomup}, baseline~\cite{rubi} and BAN~\cite{ban} respectively. We ascribe this to the following; VGQE adds a robust and visually-grounded question representation to the model without damaging its existing reasoning power.  
\section{Conclusion}
Current VQA models rely heavily on the language priors that exists in the train set, without considering the image. Such models that fail to utilize both modalities equally would likely perform poorly in real-world scenarios. We propose VGQE, a novel question-encoder that utilizes both the modalities equally and generates visually-grounded question representations. Such question representations have sufficient distinguishing power based on the visual counterpart and help the models to reduce the learning of language biases from the train set. We did extensive experiments on the bias sensitive VQA-CPv2 dataset and achieved a new state-of-the-art. The VGQE is model agnostic and can easily incorporate into existing VQA models without the need for additional manually annotated data and training. We experimented with three best performing VQA models, and deliver consistent performance improvement in all of them. Further, unlike existing bias reduction techniques, VGQE does not sacrifice the original model's performance on the standard VQAv2 benchmark. 
\section*{Supplementary material}
\subsection*{Implementation details} 
\textbf{Feature extraction:} In all our experiments, we use the fixed-sized (36) object-level image features provided by~\cite{bottomup} as $V$ ($d_v=2048$). These features are obtained from Faster-RCNN~\cite{fasterrcnn} trained on Visual Genome~\cite{genome}, with ResNet-101~\cite{resnet} as the backbone. We did not fine-tune the image features.
We use the pre-trained Glove~\cite{glove} word-embeddings (trained on ``Common Crawl") to extract the object-label and question word features ($d_w=300$). If the object-label contains two words, we take the sum of the word-embeddings of the individual words as the word-embedding vector. We use a similar question pre-processing as in~\cite{rubi}, such as lower-case transformation, removing punctuation, etc.  If a question word is not there in the Glove word embedding vocabulary, we use the word embedding of its synonym or other words that give the same meaning. 

\noindent\textbf{Model parameters:} Inside VGQE, we use a Bi-directional GRU with a hidden state dimension of $1024$ as the RNN cell. We use the fine-tuned question word embedding dimension, $d=512$. In the BLOCK fusion, we use a similar setting as in~\cite{rubi}, which consists of 15 chunks, each of rank 15, the dimension of the projection space is $1000$, and the output dimension is $2048$. 

\noindent\textbf{Dataset and training:} We use the VQA-CPv2 dataset~\cite{gvqa} to train and evaluate the language-bias reduction capacity of the models. We also use the VQAv2 dataset to train and evaluate the models on the standard VQA benchmark. For both VQA-CPv2 and VQAv2, we consider the most frequent 3000 answers as the answer vocabulary as in prior works~\cite{gvqa,overcoming,rubi} and the evaluation metric used is the VQA accuracy~\cite{vqa1}. We also use the questions from Visual Genome~\cite{genome} that matches the answer vocabulary, following the prior works~\cite{bottomup,ban}. 
We train the models using the Adamw~\cite{adamw} optimizer with $weight decay = 2*10^{-5}$ and cross-entropy loss. 
For the baseline model, we set the initial learning rate as $3.5 \times 10^{-4}$ and linearly increase it by a factor of $0.25$ till epoch 11. Then we decay the learning rate by a factor of $0.25$ with a step size of 2. For the BAN baseline, we used an initial learning rate of $2 \times 10^{-4}$ and followed the same scheduling algorithm as above. For the UpDn baseline, we fixed the learning rate as $2 \times 10^{-4}$ and used the Binary-Cross entropy loss. We use gradient clipping with a norm threshold as $0.25$. We use a batch size of 128 and \textit{dropout} value as $0.2$. For implementing all the models, we use the PyTorch~\cite{pytorch} deep-learning framework.
%
%
\bibliographystyle{splncs04}
\bibliography{egbib}
\end{document}